\documentclass{article}
\usepackage{spconf,amsmath,graphicx}
\usepackage{dirtytalk, xcolor}

\def\supdata{{\mathcal{S}}}
\def\unsupspeechdata{{\mathcal{U}^S}}
\def\unsuptextdata{{\mathcal{U}^T}}
\def\causalenc{{E_C}}
\def\noncausalenc{{E_{NC}}}
\def\causaldec{{D_C}}
\def\noncausaldec{{D_{NC}}}

\title{A COMPARISON OF SEMI-SUPERVISED LEARNING TECHNIQUES FOR STREAMING ASR AT SCALE}
\name{Cal Peyser$^{1,2}$, Michael Picheny$^{1}$, Kyunghyun Cho$^{1}$, Rohit Prabhavalkar$^{2}$, Ronny Huang$^{2}$, Tara Sainath$^{2}$}
\address{$^{1}$New York University, Center for Data Science, $^{2}$Google Inc. \\
\fontsize{9}{9}\selectfont\ttfamily\upshape
{cpeyser@google.com}}

\begin{document}
\ninept
\maketitle
\begin{abstract}
Unpaired text and audio injection have emerged as dominant methods for improving ASR performance in the absence of a large labeled corpus.   However, little guidance exists on deploying these methods to improve production ASR systems that are trained on very large supervised corpora and with realistic requirements like a constrained model size and CPU budget, streaming capability, and a rich lattice for rescoring and for downstream NLU tasks. In this work, we compare three state-of-the-art semi-supervised methods encompassing both unpaired text and audio as well as several of their combinations in a controlled setting using joint training.  We find that in our setting these methods offer many improvements beyond raw WER, including substantial gains in tail-word WER, decoder computation during inference, and lattice density.
\end{abstract}

\begin{figure*}[h]
  \centering
  \centerline{\includegraphics[width=14cm]{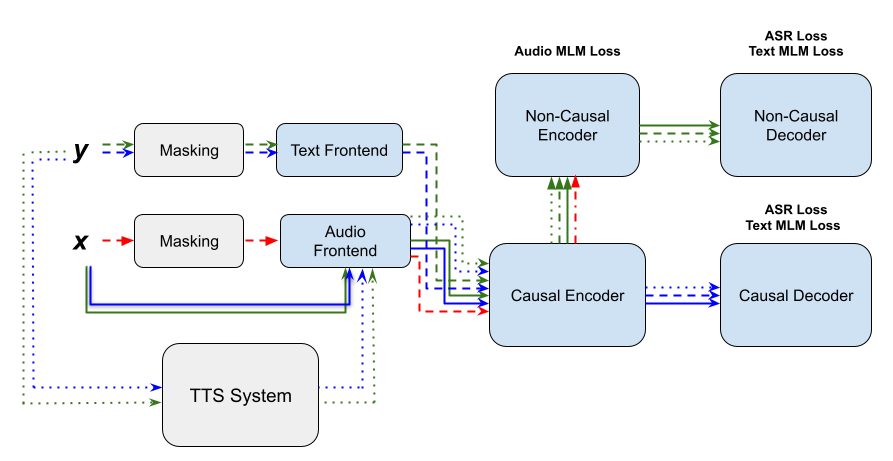}}
\caption{High-level model architecture.}
\label{fig:architecture}
\end{figure*}

\section{Introduction}
\label{sec:introduction}

Methods for learning from large-scale supervised datasets have been the primary driver of progress in speech processing from the HMM/GMM era \cite{FourResearchGroups} well into the era of deep learning \cite{RNNT, LAS, ASSAP_ASR}.  However, as the scope of ASR research has expanded into challenging settings such as low-resource languages and difficult acoustic conditions, it has become difficult to gather large-scale in-domain supervised datasets \cite{2013LowResource}.  In recent years, much of the speech community's attention has moved to alternatives to purely supervised learning.

Semi-supervised learning has emerged as a powerful paradigm for addressing contemporary problems in speech recognition \cite{LitReviewSemisupSpeech_Thesis}.  In a semi-supervised training scheme, unpaired speech and/or text examples supplement a supervised dataset to provide greater acoustic/language coverage.  A broad literature has emerged exploring various mechanisms for incorporating speech-only and text-only data into ASR training (see Section \ref{sec:related_work}).

Semi-supervised learning with both audio and text has yielded very strong results on benchmark ASR tasks, motivating interest in these methods for large-scale, production applications.  However, published results generally report on datasets smaller than industrial-scale, usually emphasizing the low-resource case in which very little supervised data is available.  Furthermore, they use large full-context architectures that do not meet realistic requirements for modern production ASR systems such as being small enough to fit on a mobile phone and capable of streaming predictions.  Finally, the literature reports almost entirely on WER improvements, with little study of measures like CPU load and lattice richness that are applicable when an ASR system acts as an individual component of an on-device system.  

In this work, we provide a comparison of several leading semi-supervised methods in a controlled setting geared towards production implementation.  Unlike previous work, we apply these methods to a state-of-the-art, 160M-parameter streaming Conformer \cite{Conformer} model that is already trained on a very large supervised corpus.  We further depart from previous work by training supervised and unsupervised tasks jointly, which is being increasingly shown to be preferable to the conventional fine-tuning approach on very large datasets \cite{JointTraining}.  We find that under these conditions, none of the studied methods improve general WER at all.  However, we report improvements in the decoder's computational load and in lattice density, as well as in several targeted WER measurements assessing performance on known categories of particularly difficult utterances.  Through this comparison and analysis, we hope to offer a more nuanced and comprehensive view of the usefulness of unpaired audio and text in industrial ASR.

The rest of our paper is structured as follows.  Section \ref{sec:related_work} summarizes the literature on the three methods under study.  Section \ref{sec:methods} presents our architecture for a streaming ASR system that supports these three methods and their combinations.  Section \ref{sec:experiments} details our datasets, experiments and evaluation criteria.  Section \ref{sec:results} presents our results, and Section \ref{sec:conclusions} concludes.
\section{Related Work}
\label{sec:related_work}
In this section, we summarize the literature surrounding the three semi-supervised learning methods that we study in this work.

\subsection{Text Injection}
Unsupervised text injection in ASR is traditionally done with language model \say{fusion}, either at inference time \cite{ShallowFusion} or training time \cite{ColdFusion, HAT}.  These methods involve the explicit separation of the model parameters into an acoustic model trained on paired data and a language model trained on unpaired text.  The improvements yielded by these methods come at the cost of the additional language model parameters at inference time.

A simultaneous line of work has sought an alternative to fusion in which unsupervised text is used to train an acoustic model directly.  One major line of work focuses on creating pseudolabels for unpaired text through synthesized audio.  This has been studied by generating a raw audio signal \cite{LRSpeech} or higher level lexical features \cite{TTS4Pretrain}.  Work adapting cycle consistency losses from machine translation have trained ASR and TTS together with a fully end-to-end objective \cite{CycleCon1, CycleCon2}. We choose TTS-based augmentation as the first method to study in this work (see Section \ref{sec:methods_tts_aug}).

Finally, a third class of methods for unpaired text injection makes use of auxiliary, text only objectives to train an ASR encoder without generating TTS pseduolabels.  Most such works have sought to train an ASR encoder to agnostically represent either audio or text, such that unpaired text is processed similarly to audio \cite{USTED, SLAM, Tang22}.  JOIST \cite{JOIST} is a recent method which does this using a masked language modeling task in the spirit of BERT \cite{BERT}.  We study JOIST in this work since it is one of the few methods that has been shown to work well together with very large supervised datasets and with on-device sized streaming models (see Section \ref{sec:methods_joist}). 

\subsection{Audio Injection}
Unsupervised audio injection is very well studied and has yielded a large literature \cite{SelfSupervisedSurvey}.  Recent work has largely built of the success of the Wav2Vec series of models \cite{Wav2Vec, Wav2Vec2_0}, which work by modeling masked segments of audio using a contrastive loss.  One line of further work investigated audio clustering to generate targets for the contrastive loss \cite{HuBERT, WavLM} while another investigated methods for computing that signal by quantizing the audio inputs \cite{vqWav2Vec}.  BEST-RQ \cite{BEiT} in particular finds that fixed random projection to a pre-initialized codebook works effectively as a quantizer.  We choose BEST-RQ as the third method to study in this work (see Section \ref{sec:methods_bestrq}).

\section{Methods}
\label{sec:methods}
In this section, we frame the problem of semi-supervised speech recognition, develop our model architecture, and specify the multi-task optimization problem that it is trained for.

\subsection{Architecture}
We are interested in the setting in which unsupervised data in both the speech and text domains is available alongside a large supervised corpus.  We denote as $(x, y) \in \supdata$ the supervised pair of a speech utterance $x$ and text label $y$ in the supervised dataset $\supdata$.  We similarly denote unsupervised speech examples as $x \in \unsupspeechdata$ and unsupervised text examples $y \in \unsuptextdata$.

We extend the cascading conformer proposed in \cite{CascadeConformer} to support semi-supervised multitask training.  To this end, we define four neural modules:

\begin{enumerate}
    \item $\causalenc$, the \say{causal} encoder, which consumes streamed audio features with no right-context.
    \item $\noncausalenc$, the \say{non-causal} encoder, which consumes the outputs of $\causalenc$ with 900ms of right-context.
    \item $\causaldec$, a decoder for the causal encoder.  During inference, this decoder may be used to generate immediate predictions as the user speaks.
    \item $\noncausaldec$, a decoder for the non-causal encoder.  During inference, this decoder may be used to revise the predictions of the causal decoder with short latency.
\end{enumerate}

Unlike \cite{CascadeConformer}, we would like our model to consume representations of either audio or text.  For this we follow JOIST, seeking mechanisms to cause the $\causalenc$ to be agnostic to the input domain.  We choose to include two neural \say{frontends}, one for audio features and one for text.  As in JOIST, we upsample text frontend outputs by repetition so that audio and text representations will be of approximately the same length.

\subsection{Tasks}
In this framework, causal and non-causal ASR are trained as they are in \cite{CascadeConformer}. In particular, for causal ASR, $x$ is processed by the audio frontend, encoded by $\causalenc$, and decoded by $\causaldec$, while non-causal ASR is processed analogously with the non-causal modules $\noncausalenc$ and $\noncausaldec$.  The model is trained end-to-end with an RNN-T \cite{RNNT} loss.  This is represented by the solid blue (causal) and solid green (non-causal) paths in Figure \ref{fig:architecture}.  For semi-supervised tasks we require different formulations.

\subsubsection{TTS Augmentation}
\label{sec:methods_tts_aug}
Using a pre-trained TTS system with frozen parameters, we generate an audio clip $\hat{x}$ corresponding to each unsupervised text segment $y \in \unsuptextdata$.  We then treat $(\hat{x}, y)$ as a supervised audio-text pair and train the causal and non-causal ASR tasks.  This is represented by the dotted blue (causal) and dotted green (causal) paths in Figure \ref{fig:architecture}.

We found that in order to achieve reasonable training speed it is important that the TTS system convert input word-pieces not into raw audio but instead into the (much shorter) sequence of acoustic features that is consumed by the audio frontend.  This is due to the fact that since the decoder of our TTS system which produces audio features is autoregressive, audio sequence length has critical implications for training speed and quickly becomes a bottleneck.

\subsubsection{JOIST}
\label{sec:methods_joist}
Following the design of JOIST in \cite{JOIST}, we pass masked unpaired text examples through a text frontend, which consists simply of a learned projection.  The results are treated identically to audio features; that is, they are passed in turn to the causal ($\causalenc$ and $\causaldec$) and non-causal ($\noncausalenc$ and $\noncausaldec$) and compared to original text sequence via an RNN-T loss.  This is represented by the dashed blue (causal) and dashed green (causal) paths in Figure \ref{fig:architecture}.

We find that it is critical for WER that JOIST consume phonemic representations of $y$, as opposed to text tokens, corroborating the findings of \cite{JOIST}.  We include a text-to-phonemes lookup in the model which processes text before masking.  The JOIST loss still operates with respect to the standard word-piece representation - that is, the JOIST loss learns to generate word pieces from a masked phoneme sequence.

\subsubsection{BEST-RQ}
\label{sec:methods_bestrq}
We model our audio injection after BEST-RQ as implemented in \cite{BEiT}.  Audio features are masked and processed by the frontend.  They are then encoded by the casual and non-causal encoders of the ASR stack.  Additionally, audio features are processed by a randomly initialized projection with frozen weights and then discretized by rounding to the nearest entry in a fixed codebook.  The encoder is then trained to predict the quantized targets inside the masked region.  This is represented by the dashed red path in Figure \ref{fig:architecture}.

\subsection{Training Scheme}
There are many approaches to multi-task semi-supervised learning, mostly focused on pretrain-finetune paradigm \cite{Wav2Vec, Wav2Vec2_0, vqWav2Vec, WavLM}.  While this methodology has achieved state of the art results on datasets such as Librispeech, we found that on our large dataset it is prone to forgetting representations learned in pretraining during finetuning, which is consistent with the findings in \cite{JointTraining} for very large training sets.  We therefore restrict our study to joint training of ASR together with the unsupervised tasks.  Note that even though joint training includes ASR, we find that it is still beneficial and convenient to initialize from a strong ASR baseline.

At each iteration during training we sample a separate batch from each dataset, $b_\supdata \in \supdata$, $b_\unsupspeechdata \in \unsupspeechdata$, and $b_\unsuptextdata \in \unsuptextdata$.  We then propagate each batch through the model, performing the preprocessing specified for TTS augmentation and JOIST on $b_\unsuptextdata$ and that specified for BEST-RQ on $b_\unsupspeechdata$.  We apply the relevant losses to each task and sum them according to specified weights.
\begin{table*}[t]
\begin{minipage}{.5\linewidth}
\centering
\begin{tabular}{||c|c|c|c|c|c||} 
 \hline
 Model & \textbf{VS} & \textbf{Noisy} & \textbf{RPN} & \textbf{R\_LM} & \textbf{C\_LM} \\ [0.5ex] 
 \hline\hline
 \textbf{E-0} & 162 & 187 & 297 & 357 & 325 \\ 
 \hline
 \textbf{E-A} & -7.2\% & -5.3\% & \textbf{-11.1\%} & \textbf{-10.1}\% & \textbf{-8.9\%} \\ 
 \hline
 \textbf{E-B} & -7.2\% & -5.3\% & -9.8\% & -8.4\% & -7.4\%\\ 
 \hline
 \textbf{E-C} & \textbf{-9.9\%} & \textbf{-6.9\%} & -6.3\% & -5.8\% & -4.9\%\\ 
 \hline
 \textbf{E-AB} & -7.2\% & -4.8\% & -6.1\% & -5.0\% & -4.0\%\\
 \hline
 \textbf{E-AC} & \textbf{-9.9\%} & -6.4\% & -10.8\% & -9.2\% & -8.3\%\\ 
 \hline
 \textbf{E-ABC} & -8.5\% & -5.9\% & -9.8\% & -8.7\% & -7.7\%\\ 
 \hline
\end{tabular}
\centering
\caption{Average Decoding States}
\label{table:average_decoding_states}
\end{minipage}
\begin{minipage}{.5\linewidth}
\centering
\begin{tabular}{||c|c|c|c|c|c||} 
 \hline
 Model & \textbf{VS} & \textbf{Noisy} & \textbf{RPN} & \textbf{R\_LM} & \textbf{C\_LM}  \\ [0.5ex] 
 \hline\hline
 \textbf{E-0} & 3.2 & 3.3 & 6.2 & 8.1 & 9.7\\ 
 \hline
 \textbf{E-A} & \textbf{+12.5\%} & \textbf{+15.2\%} & \textbf{+3.2\%} & +3.7\% & +3.1\%\\ 
 \hline
 \textbf{E-B} & \textbf{+12.5\%} & +12.1\% & \textbf{+3.2\%} & +3.7\% & +3.1\%\\ 
 \hline
 \textbf{E-C} & \textbf{+12.5\%} & +12.1\% & \textbf{+3.2\%} & +3.7\% & +3.1\%\\ 
 \hline
 \textbf{E-AB} & \textbf{+12.5\%} & +12.1\% & +1.6\% & +3.7\% & +3.1\%\\ 
 \hline
 \textbf{E-AC} & \textbf{+12.5\%} & +12.1\% & \textbf{+3.2\%} & \textbf{+4.9\%} & \textbf{+4.1\%} \\ 
 \hline
 \textbf{E-ABC} & \textbf{+12.5\%} & +12.1\% & \textbf{+3.2\%} & \textbf{+4.9\%} & +3.1\%\\ 
 \hline
\end{tabular}
\centering
\caption{Lattice Density}
\label{table:lattice_density}
\end{minipage}
\end{table*}

\section{Experiments}
\label{sec:experiments}

This section details the implementation, training, and evaluation of the architecture described above.

\subsection{Model}
Following the components in Figure \ref{fig:architecture} the architecture of our model is as follows.

The causal audio encoder $\causalenc$ consists of six conformer \cite{Conformer} layers with model dimension 2048 and eight attention heads.  The noncausal audio encoder $\noncausalenc$ adds a further nine such conformer layers.   The decoders $\causaldec$ and $\noncausaldec$ are each HAT \cite{HAT} decoders with prediction and joint networks with model dimension 640.  These four components and the audio frontend, which together make up the inference-time model, contain about 164M parameters.

The TTS system is based on Tacotron 2 \cite{Tacotron2}.  The encoder consists of three convolutions followed by a single RNN layer, while the decoder consists of a single RNN layer with attention to the encoder outputs followed by a post-net consisting of five convolutional layers.

\subsection{Training}
We train our model with a supervised dataset $\supdata$ consisting of about 4M utterances, totalling about 200k hours of speech.  We also use an unsupervised audio set $\unsupspeechdata$ of about 600M utterances and an unsupervised text set $\unsuptextdata$ of about 230B examples.
 
At timestep $t$, the audio head of our model consumes 512-dimensional features consisting of four 128-dimensional log-mel features representing the range $[t-2, t+1]$.  The log-mel features are computed at 10ms intervals and on 32ms frames.  We subsample stacked features by a factor of 3, so that each feature represents 30ms in the input.  During BEST-RQ, we a mask single span consisting of 15\% of the input features.  Text inputs are represented by a wordpiece model of size 4096.

Our baseline model is trained for 800k steps with a batch size of 2048 for each of $\supdata$, $\unsupspeechdata$, and $\unsuptextdata$.  Our semi-supervised experiments are trained for a further 35k steps, using task splits detailed in Section \ref{sec:results}.    

\subsection{Evaluation}
We evaluate our models on several test sets, seeking to measure performance under the acoustic and language conditions which are typically targeted using unsupervised data.  Our voice search test set (\textbf{VS}) is sampled from anonymized traffic to Google production services.  The \textbf{NOISY} set consists of anonymized traffic with artificial noise added.  Our remaining test sets are synthesized using a TTS system from anonymized text traffic to Google services, and are selected according to a criterion meant to target difficult language conditions.  The rare proper nouns set (\textbf{RPN}) consists of examples that contain a proper noun (as determined by a neural proper noun tagger) that occurs fewer than five times in $\supdata$.  The Rare-LM set (\textbf{R\_LM}) consists of examples containing a unigram that occurs fewer than five times in both $\supdata$ and $\unsuptextdata$, while the (\textbf{C\_LM}) consists of examples containing a unigram that occurs fewer than five times in $\supdata$ but at least 150 times in $\unsuptextdata$.  \textbf{RPN} and \textbf{C\_LM} are measure tail performance, while \textbf{C\_LM} is intended to measure the degree to which information from $\unsuptextdata$ has been incorporated into the model.
\section{Results}
\label{sec:results}
We denote JOIST with the letter \textbf{A}, TTS augmentation with \textbf{B}, and BEST-RQ with \textbf{C}.  We find the best results when each of these experiments are trained with 40\% task weighting each on causual and non-causal ASR, with the remaining 20\% split across unsupervised tasks. The weightings of the unsupervised tasks are given in Table \ref{table:task_weights}.
\begin{table}[h]
\centering
\begin{tabular}{||c||c|c|c|c|c|c||} 
 \hline
 Model & C-JOIST & NC-JOIST & TTS & BEST-RQ \\ [0.5ex] 
 \hline\hline
 \textbf{E-A} & 1/2 & 1/2 & 0 & 0 \\ 
 \hline
 \textbf{E-B} & 0 & 0 & 1 & 0 \\ 
 \hline
 \textbf{E-C} & 0 & 0 & 0 & 1 \\ 
 \hline
 \textbf{E-AB} & 1/4 & 1/4 & 1/2 & 0 \\ 
 \hline
 \textbf{E-AC} & 1/4 & 1/4 & 0 & 1/2 \\ 
 \hline
 \textbf{E-ABC} & 1/6 & 1/6 & 1/3 & 1/3 \\ 
 \hline
\end{tabular}
\centering
\caption{Task Weights. C-JOIST and NC-JOIST refer to the causal and non-causal variants.}
\label{table:task_weights}
\end{table}

We denote our baseline experiment \textbf{E-0}, which splits its weight equally between causal and non-causal supervised ASR.

We give our WER results in Table \ref{table:wer_table}.  We are unsurprised to find that given a very large supervised corpus and limited model capacity, none of our methods improve performance on the unspecialized voice search test set.  We find considerable improvement, however, under tail conditions.  JOIST consistently performs best on the acoustically clean but linguistically difficult TTS tail-word test sets, which agrees with the intuition that JOIST acts to improve the encoder's text representation.  However, JOIST in fact degrades performance on the acoustically challenging Noisy test set.  BEST-RQ seems beneficial only when combined with JOIST, where it appears to recover lost performance on noisy data while retain some of the improvements on the tail-word sets.

\begin{table}[h]
\centering
\begin{tabular}{||c|c|c|c|c|c||} 
 \hline
 Model & \textbf{VS} & \textbf{Noisy} & \textbf{RPN} & \textbf{R\_LM} & \textbf{C\_LM} \\ [0.5ex] 
 \hline\hline
 \textbf{E-0} & 6.0 & 8.2 & 21.2 & 38.3 & 55.8 \\ 
 \hline
 \textbf{E-A} & \textbf{-0.0\%} & +1.2\% & \textbf{-4.7}\% & \textbf{-5.0}\% & \textbf{-2.3}\% \\ 
 \hline
 \textbf{E-B} & \textbf{-0.0\%} & \textbf{-1.2\%} & -0.5\% & -2.1\% & -0.7\% \\ 
 \hline
 \textbf{E-C} & \textbf{-0.0\%} & +1.2\% & +0.1\% & -0.0\% & -0.4\% \\ 
 \hline
 \textbf{E-AB} & \textbf{-0.0\%} & +1.2\% & -3.8\% & -4.2\% & -2.0\% \\ 
 \hline
 \textbf{E-AC} & \textbf{-0.0\%} & -0.0\% & -2.8\% & -2.9\% & -1.2\% \\ 
 \hline
 \textbf{E-ABC} & \textbf{-0.0\%} & +2.4\% & -3.3\% & -3.7\% & -1.4\% \\ 
 \hline
\end{tabular}
\centering
\caption{Word Error Rate}
\label{table:wer_table}
\end{table}

In production systems, model performance goes beyond raw WER, since it is often not a 1-best hypothesis but rather the produced lattice that is used to generate predictions or fed directly to a downstream NLU task.  In Table \ref{table:lattice_density}, we measure the richness of the lattice by computing \say{lattice density}, which we define as the number of arcs in the lattice divided by the number of wordpieces in the ground truth.  On this measure, we find that all three methods offer considerable improvement in voice search.  For difficult utterances, we find that combinations of methods largely outperform single methods.  This agrees with the intuition that many training criteria lead to a greater diversity of plausible predictions, and invites investigation into the combination of these methods for applications like biasing or intent classification which can benefit from a rich lattice. 

Finally, since an autoregressive decoder is often a computational bottleneck in on-device systems, we seek to determine the impact of our methods on the work the decoder has to do.  In Table \ref{table:average_decoding_states}, we measure the average number of states expanded by the decoder during beam search.  We find that all three methods provide meaningful improvements over the baseline on this metric, with the best results coming from JOIST.  This suggests, unsurprisingly, that the decoder explores the fewest states when the encoder has a strong language representation.

\section{Conclusions}
\label{sec:conclusions}
In this work we apply several contemporary semi-supervised training methods to a realistic, state-of-the-art production ASR system.  We find that unlike in the conventional setting, with a large full-context model and only a small amount of supervised data, these methods do not offer improvement on unspecialized WER.  We demonstrate, however, that these techniques nevertheless offer meaningful utility for tail-condition performance, lattice density, and decoder computational load.  We believe that these results motivate a broader perspective on semi-supervised training in its application to industrial ASR.

\bibliographystyle{IEEEbib}
\bibliography{strings,refs}

\begin{thebibliography}{10}

\bibitem{FourResearchGroups}
Geoffrey Hinton, Li~Deng, Dong Yu, George~E. Dahl, Abdel-rahman Mohamed,
  Navdeep Jaitly, Andrew Senior, Vincent Vanhoucke, Patrick Nguyen, Tara~N.
  Sainath, and Brian Kingsbury,
\newblock ``Deep neural networks for acoustic modeling in speech recognition:
  The shared views of four research groups,''
\newblock {\em IEEE Signal Processing Magazine}, vol. 29, no. 6, pp. 82--97,
  2012.

\bibitem{RNNT}
Alex Graves,
\newblock ``Sequence transduction with recurrent neural networks,''
\newblock in {\em International Conference on Machine Learning (ICML)}, 2012.

\bibitem{LAS}
William Chan, Navdeep Jaitly, Quoc Le, and Oriol Vinyals,
\newblock ``Listen, attend and spell: A neural network for large vocabulary
  conversational speech recognition,''
\newblock in {\em 2016 IEEE International Conference on Acoustics, Speech and
  Signal Processing (ICASSP)}, 2016, pp. 4960--4964.

\bibitem{ASSAP_ASR}
Jing Pan, Joshua Shapiro, Jeremy Wohlwend, Kyu Han, Tao Lei, and Tao Ma,
\newblock ``Asapp-asr: Multistream cnn and self-attentive sru for sota speech
  recognition,''
\newblock 10 2020, pp. 16--20.

\bibitem{2013LowResource}
Samuel Thomas, Michael~L. Seltzer, Kenneth Church, and Hynek Hermansky,
\newblock ``Deep neural network features and semi-supervised training for low
  resource speech recognition,''
\newblock in {\em 2013 IEEE International Conference on Acoustics, Speech and
  Signal Processing}, 2013, pp. 6704--6708.

\bibitem{LitReviewSemisupSpeech_Thesis}
Jennifer Drexler,
\newblock {\em Deep unsupervised learning from speech},
\newblock Ph.D. thesis, 01 2016.

\bibitem{Conformer}
Anmol Gulati, James Qin, Chung-Cheng Chiu, Niki Parmar, Yu~Zhang, Jiahui Yu,
  Wei Han, Shibo Wang, Zhengdong Zhang, Yonghui Wu, and Ruoming Pang,
\newblock ``Conformer: Convolution-augmented transformer for speech
  recognition,''
\newblock in {\em INTERSPEECH}, 2020.

\bibitem{JointTraining}
Junwen Bai, Bo~Li, Yu~Zhang, Ankur Bapna, Nikhil Siddhartha, Khe~Chai Sim, and
  Tara~N. Sainath,
\newblock ``Joint unsupervised and supervised training for multilingual
  {ASR},''
\newblock in {\em International Conference on Acoustics, Speech, and Signal
  Processing (ICASSP)}, 2021.

\bibitem{ShallowFusion}
{\c{C}}aglar G{\"{u}}l{\c{c}}ehre, Orhan Firat, Kelvin Xu, Kyunghyun Cho,
  Lo{\"{\i}}c Barrault, Huei{-}Chi Lin, Fethi Bougares, Holger Schwenk, and
  Yoshua Bengio,
\newblock ``On using monolingual corpora in neural machine translation,''
\newblock {\em CoRR}, vol. abs/1503.03535, 2015.

\bibitem{ColdFusion}
Anuroop Sriram, Heewoo Jun, Sanjeev Satheesh, and Adam Coates,
\newblock ``Cold fusion: Training seq2seq models together with language
  models,''
\newblock {\em CoRR}, vol. abs/1708.06426, 2017.

\bibitem{HAT}
Ehsan Variani, David Rybach, Cyril Allauzen, and Michael Riley,
\newblock ``Hybrid autoregressive transducer (hat),''
\newblock in {\em International Conference on Acoustics, Speech, and Signal
  Processing (ICASSP)}, 2020.

\bibitem{LRSpeech}
Jin Xu, Xu~Tan, Yi~Ren, Tao Qin, Jian Li, Sheng Zhao, and Tie-Yan Liu,
\newblock ``Lrspeech: Extremely low-resource speech synthesis and
  recognition,''
\newblock in {\em ACM SIGKDD International Conference on Knowledge Discovery
  and Data Mining}, 2020.

\bibitem{TTS4Pretrain}
Zhehuai Chen, Yu~Zhang, Andrew Rosenberg, Bhuvana Ramabhadran, Gary Wang, and
  Pedro~J. Moreno,
\newblock ``Injecting text in self-supervised speech pretraining,''
\newblock in {\em IEEE Automatic Speech Recognition and Understanding Workshop
  (ASRU)}, 2021.

\bibitem{CycleCon1}
Takaaki Hori, Ram{\'{o}}n~Fernandez Astudillo, Tomoki Hayashi, Yu~Zhang, Shinji
  Watanabe, and Jonathan~Le Roux,
\newblock ``Cycle-consistency training for end-to-end speech recognition,''
\newblock in {\em International Conference on Acoustics, Speech, and Signal
  Processing (ICASSP)}, 2019.

\bibitem{CycleCon2}
Murali~Karthick Baskar, Shinji Watanabe, Ramon Astudillo, Takaaki Hori, Lukáš
  Burget, and Jan Černocký,
\newblock ``Semi-supervised sequence-to-sequence asr using unpaired speech and
  text,''
\newblock in {\em INTERSPEECH}, 2019.

\bibitem{USTED}
Bolaji Yusuf, Ankur Gandhe, and Alex Sokolov,
\newblock ``Usted: Improving asr with a unified speech and text
  encoder-decoder,''
\newblock in {\em International Conference on Acoustics, Speech, and Signal
  Processing (ICASSP)}, 2022.

\bibitem{SLAM}
Ankur Bapna, Yu{-}An Chung, Nan Wu, Anmol Gulati, Ye~Jia, Jonathan~H. Clark,
  Melvin Johnson, Jason Riesa, Alexis Conneau, and Yu~Zhang,
\newblock ``{SLAM:} {A} unified encoder for speech and language modeling via
  speech-text joint pre-training,''
\newblock in {\em Annual Meeting of the Association for Computational
  Linguistics (ACL)}, 2021.

\bibitem{Tang22}
Yun Tang, Hongyu Gong, Ning Dong, Changhan Wang, Wei-Ning Hsu, Jiatao Gu,
  Alexei Baevski, Xian Li, Abdelrahman Mohamed, Michael Auli, and Juan Pino,
\newblock ``Unified speech-text pre-training for speech translation and
  recognition,''
\newblock in {\em Annual Meeting of the Association for Computational
  Linguistics (ACL)}, 2022.

\bibitem{JOIST}
Tara~N. Sainath, Rohit Prabhavalkar, A.~Bapna, Y.~Zu, Z.~Huo, Z.~Chen, B.~Li,
  W.~Wang, and T.~Strohman,
\newblock ``Joist: A joint speech and text streaming model for asr,''
\newblock in {\em IEEE Spoken Language Technology Workshop (SLT)}, 2022.

\bibitem{BERT}
Jacob Devlin, Ming{-}Wei Chang, Kenton Lee, and Kristina Toutanova,
\newblock ``{BERT:} pre-training of deep bidirectional transformers for
  language understanding,''
\newblock {\em CoRR}, vol. abs/1810.04805, 2018.

\bibitem{SelfSupervisedSurvey}
Abdelrahman Mohamed, Hung-yi Lee, Lasse Borgholt, Jakob~D. Havtorn, Joakim
  Edin, Christian Igel, Katrin Kirchhoff, Shang-Wen Li, Karen Livescu, Lars
  Maaløe, Tara~N. Sainath, and Shinji Watanabe,
\newblock ``Self-supervised speech representation learning: A review,''
\newblock {\em IEEE Journal of Selected Topics in Signal Processing}, vol. 16,
  no. 6, pp. 1179--1210, 2022.

\bibitem{Wav2Vec}
Steffen Schneider, Alexei Baevski, Ronan Collobert, and Michael Auli,
\newblock ``wav2vec: Unsupervised pre-training for speech recognition,''
\newblock {\em CoRR}, vol. abs/1904.05862, 2019.

\bibitem{Wav2Vec2_0}
Alexei Baevski, Henry Zhou, Abdelrahman Mohamed, and Michael Auli,
\newblock ``wav2vec 2.0: {A} framework for self-supervised learning of speech
  representations,''
\newblock {\em CoRR}, vol. abs/2006.11477, 2020.

\bibitem{HuBERT}
Wei{-}Ning Hsu, Benjamin Bolte, Yao{-}Hung~Hubert Tsai, Kushal Lakhotia, Ruslan
  Salakhutdinov, and Abdelrahman Mohamed,
\newblock ``Hubert: Self-supervised speech representation learning by masked
  prediction of hidden units,''
\newblock {\em CoRR}, vol. abs/2106.07447, 2021.

\bibitem{WavLM}
Sanyuan Chen, Chengyi Wang, Zhengyang Chen, Yu~Wu, Shujie Liu, Zhuo Chen, Jinyu
  Li, Naoyuki Kanda, Takuya Yoshioka, Xiong Xiao, Jian Wu, Long Zhou, Shuo Ren,
  Yanmin Qian, Yao Qian, Jian Wu, Michael Zeng, and Furu Wei,
\newblock ``Wavlm: Large-scale self-supervised pre-training for full stack
  speech processing,''
\newblock {\em CoRR}, vol. abs/2110.13900, 2021.

\bibitem{vqWav2Vec}
Alexei Baevski, Steffen Schneider, and Michael Auli,
\newblock ``vq-wav2vec: Self-supervised learning of discrete speech
  representations,''
\newblock in {\em International Conference on Learning Representations (ICLR)},
  2019.

\bibitem{BEiT}
Chung{-}Cheng Chiu, James Qin, Yu~Zhang, Jiahui Yu, and Yonghui Wu,
\newblock ``Self-supervised learning with random-projection quantizer for
  speech recognition,''
\newblock in {\em International Conference on Machine Learning (ICML)}, 2022.

\bibitem{CascadeConformer}
Arun Narayanan, Tara~N. Sainath, Ruoming Pang, Jiahui Yu, Chung-Cheng Chiu,
  Rohit Prabhavalkar, Ehsan Variani, and Trevor Strohman,
\newblock ``Cascaded encoders for unifying streaming and non-streaming asr,''
\newblock in {\em International Conference on Acoustics, Speech, and Signal
  Processing (ICASSP)}, 2021.

\bibitem{Tacotron2}
Jonathan Shen, Ruoming Pang, Ron~J. Weiss, Mike Schuster, Navdeep Jaitly,
  Zongheng Yang, Zhifeng Chen, Yu~Zhang, Yuxuan Wang, R.~J. Skerry{-}Ryan,
  Rif~A. Saurous, Yannis Agiomyrgiannakis, and Yonghui Wu,
\newblock ``Natural {TTS} synthesis by conditioning wavenet on mel spectrogram
  predictions,''
\newblock in {\em International Conference on Acoustics, Speech, and Signal
  Processing (ICASSP)}, 2018.

\end{thebibliography}

\end{document}